%% file: main.tex
\pdfoutput=1
\documentclass[10pt,twocolumn,letterpaper]{article}

\usepackage[applications,pagenumbers]{wacv}

\input{preamble}

\definecolor{wacvblue}{rgb}{0.21,0.49,0.74}
\usepackage[breaklinks,colorlinks,allcolors=wacvblue]{hyperref}
\usepackage[capitalize]{cleveref}

\graphicspath{{figs/}}

\def\wacvPaperID{309}
\def\confName{WACV}
\def\confYear{2027}

\title{FALCON: FPGA-Accelerated Latency-Certifiable Occupancy Networks\\for Camera-Based Autonomous Driving}

\author{Haohua Que\thanks{Equal contribution.} \qquad Mingkai Liu\footnotemark[1] \qquad Handong Yao\thanks{Corresponding author.}\\
College of Engineering, University of Georgia, Athens, Georgia, USA\\
{\tt\small hq10606@uga.edu \qquad MingkaiKLEIN@outlook.com \qquad handong.yao@uga.edu}}

\makeatletter
\patchcmd{\@maketitle}{\vspace*{24pt}}{\vspace*{12pt}}{}{}
\makeatother

\begin{document}
\maketitle

\input{01-abstract-intro}
\input{02-related-work}
\input{03-method}
\input{04-experiments}
\input{05-discussion}

{\small
\setlength{\bibsep}{0pt}
\bibliographystyle{ieeenat_fullname}
\bibliography{references}
}

\end{document}

%% file: preamble.tex
\usepackage{array}
\newcolumntype{L}[1]{>{\raggedright\arraybackslash}p{#1}}

%% file: 01-abstract-intro.tex
\begin{abstract}
A perception system that must brake for a pedestrian does not need a low average latency. It must deliver an
answer before the deadline on \emph{every} frame, the worst case included. Commodity edge accelerators
optimize the mean through dynamic scheduling, OS jitter, thermal throttling, and contention, so their
per-frame latency can be measured but not bounded. We build FALCON, a camera-based 3D occupancy front-end on
the AMD Kria KR260, an embedded FPGA drawing $4.3$\,W idle and up to $6.6$\,W under inference, and make the
latency guarantee \textbf{layered and explicit}: programmable-logic (PL) compute is cycle-bounded
analytically; inference is a measured envelope over $100$k-frame traces; and a watchdog enforces
\emph{output delivery}, publishing the previous output flagged stale when a frame is late. The deadline runs
use a fixed-work proxy of six ResNet-50 DPU inferences per frame, not the full occupancy network: across
three $100$k-frame enforced runs, all $300$k proxy frames delivered an output within the $100$\,ms deadline
plus at most $10.2$\,ms, the worst of the three runs. The full six-camera FlashOcc pipeline composes to about
$1.3$\,s per frame on the board and is not yet a $10$\,Hz system. The two operations the DPU cannot run, a
view-transform \emph{gather} and a \emph{bilinear resize}, become fixed-trip-count PL IPs numerically matched
to their software references. On the full Occ3D-nuScenes validation set the board reaches mIoU
$\mathbf{29.88}$, $92.8\%$ of an FP32 baseline the DPU cannot run. On a ResNet-50 probe over $100$k frames
the KR260 tail stays within $1.5\times$ its median ($2.0\times$ for the six-inference proxy), where Jetson
Orin reaches $4.4\times$ and the worst datacenter GPU run (NVIDIA L4) $23.6\times$. The deployment recipe, HLS
sources, and raw timing traces are released.
\end{abstract}

\section{Introduction}
\label{sec:intro}

\begin{figure}[t!]
\centering
\includegraphics[width=\linewidth]{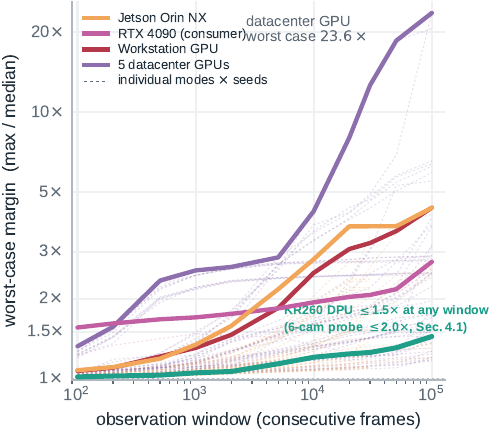}
\caption{Tail ratio (max\,/\,median) versus observation window on $100$k-frame traces of a ResNet-50 probe
(INT8 on KR260 and Orin, FP16 on GPUs). Bold lines: each platform's worst run; dashed: power-mode and seed
spread. The KR260 DPU tail grows slowly because accelerator compute is fixed by cycle count; rare host-path
spikes (up to $2.0\times$ on the six-inference proxy) are left to the delivery watchdog. Orin NX reaches $4.4\times$ and the worst datacenter run (NVIDIA L4,
no co-tenant, shared cluster node) $23.6\times$.}
\label{fig:window}
\vspace{-15pt}
\end{figure}

A self-driving stack that brakes for a pedestrian needs perception before the braking deadline on
\textbf{every} frame, not just low average latency. Occupancy prediction, a dense 3D map of free and occupied
space~\cite{occ3d,surroundocc,occnet}, is becoming that safety-bearing layer for vulnerable road users (VRUs:
pedestrians, cyclists, motorcyclists). Yet occupancy methods, from lift-splat models~\cite{lss,bevdet,flashocc}
to sparse and set-prediction models~\cite{sparseocc,opus}, are judged mainly by mIoU and sometimes mean FPS.
Mean throughput is the wrong safety currency: on a late frame, the pedestrian is seen too late.

A safety case uses worst-case execution time (WCET), a per-frame bound checked against the deadline, and
ISO~26262~\cite{iso26262} asks for that evidence rather than an average. GPUs cannot provide it: datacenter
GPUs and automotive edge SoCs optimize the mean, while dynamic scheduling, driver and OS jitter, thermal
throttling, and contention move the tail~\cite{dnn-timevar}. On a ResNet-50 probe, Jetson Orin's worst case keeps
climbing with the observation window and swings from $1.1\times$ to $4.4\times$ across power modes
and seeds, unlimited power included (\cref{fig:window}). A GPU
latency can be measured but not bounded. A bound needs both a statically scheduled accelerator whose cycle
count is fixed~\cite{finn,restuccia-rtss21} and a model with \emph{input-invariant} work. This is independent
of accuracy: recent occupancy models differ by about $25\times$ in determinism, with an input-dependent sparse
model at $27.5\%$ coefficient of variation (CV) and a fixed-query sparse model at $1.1\%$.

This paper builds such a front end on the AMD Kria KR260, an embedded FPGA at $4.3$\,W idle and up to $6.6$\,W under inference. We choose
FlashOcc~\cite{flashocc} over sparse SOTA because it is dense, fixed-compute, and DPU-compilable
(\cref{sec:exp-taxonomy}). Two stages the DPU cannot run, the lift-splat view transform~\cite{lss,bevpoolv2}
and BEV-neck bilinear resize, become custom HLS IPs: with a fixed camera rig their indices and weights are
static constants, so both are fixed-trip-count operators whose compute WCET is a cycle count. Accuracy survives a DPU-compatible INT8 recipe with calibration-time
activation clipping and no QAT~\cite{dpuczdx8g,vitisai,qdbev}: on full Occ3D-nuScenes~\cite{occ3d,nuscenes}
validation the board reaches mIoU $\mathbf{29.88}$, $1.55$ below the INT8 reference matched to its deployed
configuration and $92.8\%$ of an FP32 baseline the DPU cannot run (\cref{sec:int8-accuracy}).

Only PL compute ($11.6$\,ms per frame) is bounded analytically; DPU
stages, DDR streaming, and the host path are measured envelopes, and the watchdog guarantees that \emph{an}
output, possibly stale, is published within the deadline plus a measured envelope. The $100$\,ms deadline runs use a fixed-work proxy of six
ResNet-50 DPU inferences per frame, while the full pipeline composes to about $1.3$\,s (\cref{tab:budget}).
Our contribution is threefold:
\begin{enumerate}
\item \textbf{An evaluation axis for safety-critical occupancy.} Latency-determinability: input-dependent
computation, not sparsity, drives tails, and we decompose non-determinism into architecture, platform
sharing, and host path (\cref{sec:exp-legs}).
\item \textbf{Two fixed-trip-count PL IPs that close the DPU gaps.} Lift-splat gather and BEV-neck bilinear
resize as custom HLS IPs with cycle-bounded, input-invariant compute, numerically matched to their software
references (\cref{sec:discussion}).
\item \textbf{A quantized board deployment and a measured, enforced deadline.} Every stage deployed and
measured on the board (\cref{tab:accuracy,tab:budget}); deadline behavior characterized at $100$k-frame
scale on the fixed-work proxy against GPU and Orin (\cref{fig:deadline}); delivery enforced, with its
envelope and its fallback cost measured.
\end{enumerate}

%% file: 02-related-work.tex
\section{Related Work}\label{sec:related}

\begin{figure*}[t]
\centering
\includegraphics[width=\textwidth]{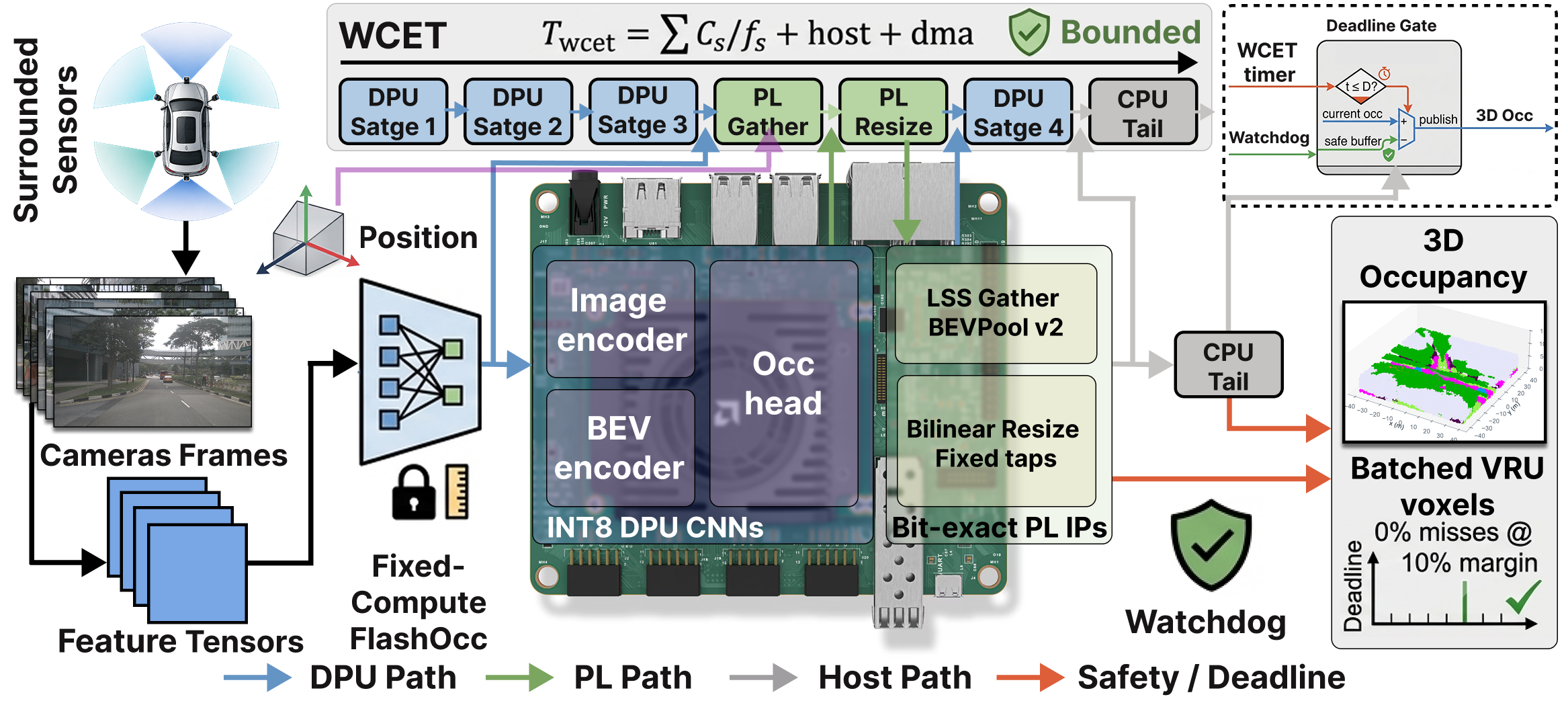}
\caption{System overview: six cameras feed fixed-compute FlashOcc; DPU stages run dense INT8 convolutions; custom PL IPs implement gather and resize; the watchdog enforces the deadline before the 3D occupancy output.}
\label{fig:datapath}
\vspace{-10pt}
\end{figure*}

Camera-only 3D occupancy descends from Lift-Splat-Shoot~\cite{lss}. BEVDet~\cite{bevdet} and BEVDepth~\cite{bevdepth} established the deployment-friendly LSS pipeline; SimpleOccupancy~\cite{simpleoccupancy}, OccNet~\cite{occnet}, TPVFormer~\cite{tpvformer}, VoxFormer~\cite{voxformer}, OccFormer~\cite{occformer}, and COTR~\cite{cotr} explore compact, dense, plane-based, sparse, and transformer-style occupancy representations. RenderOcc uses 2D rendering supervision~\cite{renderocc}; BEVFormer~\cite{bevformer}, SurroundOcc~\cite{surroundocc}, FB-OCC~\cite{fbocc}, OpenOccupancy~\cite{openoccupancy}, and GaussianFormer~\cite{gaussianformer} buy accuracy at higher cost; SparseOcc~\cite{sparseocc} and OPUS~\cite{opus} cut dense-voxel cost. We deploy FlashOcc~\cite{flashocc}, which keeps features in 2D BEV and recovers height with a channel-to-height plug-in. These works optimize mIoU and sometimes mean FPS, not the per-frame latency determinism that decides deadline feasibility (\cref{sec:experiments}).

Low-bit BEV/occupancy quantization is mostly GPU/TensorRT-targeted: QD-BEV~\cite{qdbev} and PTQAT~\cite{ptqat} use QAT, BEVFormer\_tensorrt~\cite{bevformer-trt} uses INT8 PTQ, and all rest on general methods such as AdaRound~\cite{adaround}, BRECQ~\cite{brecq}, SmoothQuant~\cite{smoothquant}, HAQ~\cite{haq}, and clipping-based activation quantization~\cite{pact}. The AMD/Xilinx DPU~\cite{dpuczdx8g} with vai\_q\_pytorch~\cite{vitisai} is more constrained: per-tensor power-of-two scales and min-max calibration only. We therefore use fixed activation clipping to stay inside the supported DPU path; prior quantized occupancy work rarely reports worst-case timing.

On the deployment-operator side, BEVPoolv2~\cite{bevpoolv2} precomputes ranks and intervals for fast GPU LSS pooling, FastOcc~\cite{fastocc} shares 2D BEV and perspective-view features, and TensorRT ports rely on custom plugins~\cite{bevformer-trt}. Their kernels remain GPU software with scheduler jitter, whereas our PL gather and resize turn fixed camera tables into on-chip memories with static trip counts and reported FPGA WCET.

For embedded-FPGA perception, the argument is timing: GPU per-frame latency is hard to bound~\cite{dnn-timevar}, while ISO 26262~\cite{iso26262} values worst-case evidence over average throughput. Restuccia and Biondi~\cite{restuccia-rtss21} analyze the Zynq UltraScale+ DPU, and FINN/FINN-R~\cite{finn,finn-r} with PointPillars-on-FINN~\cite{pp-finn} is the closest embedded 3D precedent. Yet PointPillars-on-FINN is LiDAR-based and leaves scatter on ARM, while DPU WCET work treats the accelerator as a generic CNN. No prior work gives a camera-to-occupancy datapath both cycle-bounded view operators and an enforced delivery layer on programmable logic, including the two operations the DPU cannot run.

%% file: 03-method.tex
\section{An Occupancy Front End with Certifiable Delivery}
\label{sec:method}

\subsection{From latency determinability to a hardware split}
\label{sec:method-platform}
\label{sec:method-split}
The method follows from three requirements: the model must have input-invariant work; any operation the accelerator cannot run must become a fixed-trip-count hardware operator, not move to a jittery CPU path; and the INT8 deployment must preserve the quantized algorithm with a DPU-compatible calibration recipe~\cite{jacob2018integer} rather than training-heavy QAT~\cite{lsq}.

We target the AMD Kria KR260~\cite{kria_kr260} (K26 Zynq UltraScale+ SOM, B4096 DPU, 4\,GB DDR4, measured $4.3$\,W idle to $8.7$\,W peak) with the DPUCZDX8G accelerator~\cite{dpuczdx8g}, and deploy FlashOcc-R50~\cite{flashocc} on Occ3D-nuScenes~\cite{occ3d,nuscenes}. FlashOcc keeps features in 2D BEV and recovers height with a channel-to-height plug-in, a zero-FLOP reshape of BEV channels into height bins. It is \textbf{fixed-compute}, one forward pass with input-independent FLOPs, where sparse, transformer, rendering, and efficiency-oriented designs reduce average work instead~\cite{voxformer,occformer,tpvformer,renderocc,fastocc}; a deadline bound requires fixed per-frame work. For a pipeline with hardware stages $s$, a certifiable deadline uses a static timing bound, not a measured average~\cite{wcet-survey}:
\begin{equation}
T_{\mathrm{wcet}} =
T_{\mathrm{host}}^{\mathrm{iso}} +
\sum_s \frac{C_s}{f_s} + T_{\mathrm{dma}},
\qquad
C_s(x) = C_s \;\; \forall x .
\label{eq:wcet}
\end{equation}
Here $C_s$ is the stage cycle count, $f_s$ its clock, and $x$ the scene. The important condition is $C_s(x)=C_s$: values may change, but the amount of work cannot. In this work only the $C_s/f_s$ terms are analytic; $T_{\mathrm{host}}^{\mathrm{iso}}$ and $T_{\mathrm{dma}}$ are measured envelopes (\cref{sec:exp-ips}), and the deadline guarantee that does not depend on either is enforced at the delivery layer (\cref{eq:watchdog}).

For our split, the deadline margin for a selected deadline $D$ is
\begin{equation}
\begin{aligned}
M(D) &= D - T_{\mathrm{pipe}},\\
T_{\mathrm{pipe}} &=
T_{\mathrm{host}}^{\mathrm{iso}}
+ \sum_{k \in \mathcal{D}}\frac{C^{\mathrm{dpu}}_k}{f_{\mathrm{dpu}}}
+ \frac{C_{\mathrm{gather}}+C_{\mathrm{resize}}}{f_{\mathrm{pl}}}
+ T_{\mathrm{dma}} .
\end{aligned}
\label{eq:deadline-margin}
\end{equation}
where $\mathcal{D}$ is the set of compiled DPU subgraphs. A viable operating point needs $M(D)\geq 0$; with measured envelope terms this is a design target, and the certified property remains delivery (\cref{eq:watchdog}).

The split in \cref{fig:datapath} enforces this condition. Dense convolutions run as INT8 DPU subgraphs: image trunk, neck, BEV encoder, and convolutional occupancy head. Two FlashOcc operations are not DPU primitives: the LSS view transform~\cite{lss,bevpoolv2} and BEV-neck bilinear resize. Sending them to ARM would put OS jitter back onto the critical path, so both move to programmable logic (PL) as custom HLS operators~\cite{hls4ml,fpdnn}. The final softplus-plus-linear predictor over 40\,000 BEV cells is fixed-cost but not small ($22$\,GFLOP per frame); we measure it on the CPU and prototype it on the DPU in \cref{sec:exp-payoff}.

\subsection{Static-index gather operators}
\label{sec:method-static}
\label{sec:method-gather}
\label{sec:method-resize}
Both non-DPU operations are irregular only because of geometry, and the geometry is fixed. The LSS projection tables are functions of intrinsics, extrinsics, and preprocessing; bilinear coordinates and weights are functions of tensor shapes. Once the rig is fixed, addresses, weights, and loop trip counts are constants. The two operators therefore become static-index gathers.

For the LSS gather, with BEV pillar $p$ and channel $c$, the hardware computes
\begin{equation}
B_{p,c} = \sum_{i \in \mathcal{I}_p}
D_{r_i^d} \, F_{r_i^f,c},
\qquad
\mathcal{I}_p,\; r_i^d,\; r_i^f \;\; \text{fixed by calibration}.
\label{eq:lss-gather}
\end{equation}
The interval $\mathcal{I}_p$ and ranks are static for the rig, so the total trip count $N_{\mathrm{points}}$ is fixed. Segmented accumulation writes each BEV cell once, avoiding scatter hazards. The HLS kernel unrolls $C=64$ channels over a 512-bit AXI bus, uses INT8 inputs with a wide fixed-point accumulator, and writes INT16 BEV features in the deployed path to preserve occupied-cell fidelity.
Equivalently, the compiled table is
\begin{equation}
\mathcal{T}_{\mathrm{lss}}=
\{(p_i,r_i^d,r_i^f)\}_{i=1}^{N_{\mathrm{points}}},
\qquad
C_{\mathrm{gather}}=
C_0^{\mathrm{g}}+
N_{\mathrm{points}}\left\lceil \frac{C}{P_c}\right\rceil ,
\label{eq:gather-wcet}
\end{equation}
where $P_c$ is the channel parallelism and $C_0^{\mathrm{g}}$ is pipeline fill and drain overhead. Since $\mathcal{T}_{\mathrm{lss}}$ is fixed after calibration, $C_{\mathrm{gather}}$ is independent of scene content.

For bilinear resize, at output location $(u,v)$,
\begin{equation}
Y_{u,v,c} =
\sum_{a,b \in \{0,1\}}
w^y_{u,a} w^x_{v,b}
X_{y_{u,a},x_{v,b},c},
\label{eq:bilinear}
\end{equation}
where indices $y_{u,a},x_{v,b}$ and weights $w^y,w^x$ are fixed by the input and output shapes. We implement this as a separable two-pass gather with row fusion, avoiding the full-plane intermediate that overflows BRAM. Together, \cref{eq:lss-gather,eq:bilinear} cover exactly the two non-DPU operations; \cref{tab:hlsips} reports exactness, utilization, and WCET.
For input shape $(H_i,W_i,C)$ and output shape $(H_o,W_o,C)$, the resize IP has
\begin{equation}
C_{\mathrm{resize}}=
C_0^{\mathrm{r}}+
\left(H_o W_i+H_o W_o\right)
\left\lceil \frac{C}{P_c}\right\rceil ,
\label{eq:resize-wcet}
\end{equation}
again up to fixed pipeline overhead. The two terms correspond to the fused separable passes, so the bound is a function of tensor shapes and hardware parallelism only.

\begin{figure*}[t]
\centering
\includegraphics[width=\textwidth]{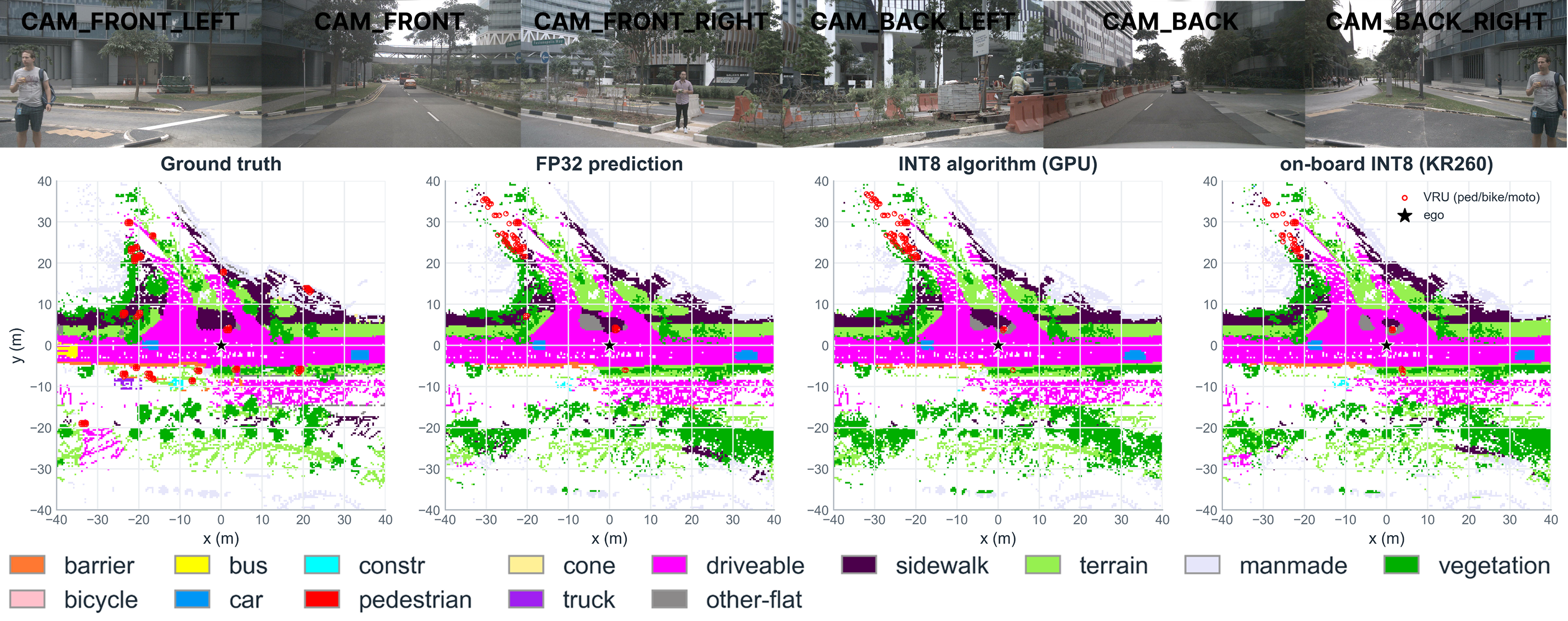}
\caption{Qualitative deployment example: ground truth, FP32, GPU INT8, and KR260 INT8 outputs show the board tracking the INT8 reference and preserving visible VRU structure.}
\label{fig:occ}
\vspace{-14pt}
\end{figure*}

\subsection{Deadline path and INT8 recipe}
\label{sec:method-determinism}
\label{sec:int8}
\label{sec:int8-trunk}
\label{sec:int8-outlier}
The timing path is governed in three layers: PL compute is cycle-bounded; inference latency is a measured envelope; delivery is enforced by a watchdog that publishes the previous ego-warped occupancy, flagged stale, when a frame would miss the deadline. Enforcement is implemented and measured on the board: fallback preparation costs $0.2$\,ms on the A53, and every frame yields an output at the deadline plus an enforcement envelope of $0.3$ to $1.9$\,ms typical, $7.6$ to $10.2$\,ms worst across the $421$ events of three enforced $100$k-frame runs (\cref{sec:exp-payoff}), a host-thread wake latency that a PL-side timer, not implemented here, would remove. The fallback is conservative only in that known occupied space is retained under ego motion; a VRU first appearing during the missed frame is invisible to it, which is why the output carries the stale flag and why the fallback's cost is measured directly (\cref{sec:exp-robust}). The host uses core dedication rather than RT priority, the stable policy for the polling DPU runner (\cref{sec:exp-legs}). For frame $t$,
\begin{equation}
\begin{aligned}
O_t^{\mathrm{out}}=
\begin{cases}
\hat O_t, & T_t \leq D,\\
W_{\Delta E_t}(O_{t-1}^{\mathrm{out}}), & T_t>D,
\end{cases}
\\
r_{\mathrm{miss}}(D)&=\frac{1}{N}\sum_{t=1}^{N}\mathbf{1}[T_t>D].
\end{aligned}
\label{eq:watchdog}
\end{equation}
Here $\hat O_t$ is the current occupancy prediction and $W_{\Delta E_t}$ warps by the ego-pose change $\Delta E_t$ accumulated since the last delivered frame (implemented as a precomputed constant-velocity shift, prepared during the previous frame's slack), and $r_{\mathrm{miss}}$ is the deadline-miss rate used in \cref{fig:payoff}.

The DPU path uses Vitis AI per-tensor power-of-two INT8~\cite{vitisai,dpuczdx8g}. The image trunk verifies the toolchain: the on-board DPU output is identical to the quantizer simulation (\cref{sec:exp-int8}). The BEV stage needs one calibration adjustment. Naive PTQ fails because LSS creates a \textbf{34$\times$ activation outlier} at the first BEV layer (\texttt{max}=206.6 vs p99.9=6.0). Under min-max scaling, most activations receive only about five effective INT8 levels. Instead of QAT-style recovery used by BEV quantizers~\cite{qdbev}, we keep the DPU post-training quantization path~\cite{dfq,zeroq,ptq4vit} and clip the activation range before quantization:
\begin{equation}
\hat a = \min(\max(a,0), C), \qquad C=32,
\label{eq:clamp}
\end{equation}
with $C=32$ chosen from calibration statistics. This is ordinary activation clipping applied at calibration time. The clipped values are saturated and near-uninformative, so the operation has negligible FP32 cost, removes the scale outlier from min-max calibration, and requires \textbf{no QAT}. The accuracy impact is measured in \cref{sec:int8-accuracy}.

The resulting DPU input is quantized with a power-of-two scale,
\begin{equation}
q(a)=
\operatorname{sat}_{[-128,127]}
\left(\operatorname{round}\left(\frac{\hat a}{2^k}\right)\right),
\qquad k\in \mathbb{Z},
\label{eq:dpu-quant}
\end{equation}
which matches the Vitis AI constraint and keeps the deployed model aligned with the simulated INT8 reference.

%% file: 04-experiments.tex
\section{Experiments}
\label{sec:experiments}

\subsection{Latency determinability}
\label{sec:exp-determinism}
\label{sec:exp-taxonomy}
\label{sec:exp-legs}

We report host-visible per-frame latency using one-frame submit-and-wait execution. The sweep covers a dedicated KR260 DPU, a dedicated RTX PRO 6000, a shared A100, a Jetson Orin NX (TensorRT 10.3 INT8, 10 to 15\,W), and three occupancy models on the shared A100: FlashOcc, SparseOcc, and OPUS. Per-frame statistics exclude a single warmup frame.

\begin{figure*}[t]
\centering
\includegraphics[width=0.595\textwidth]{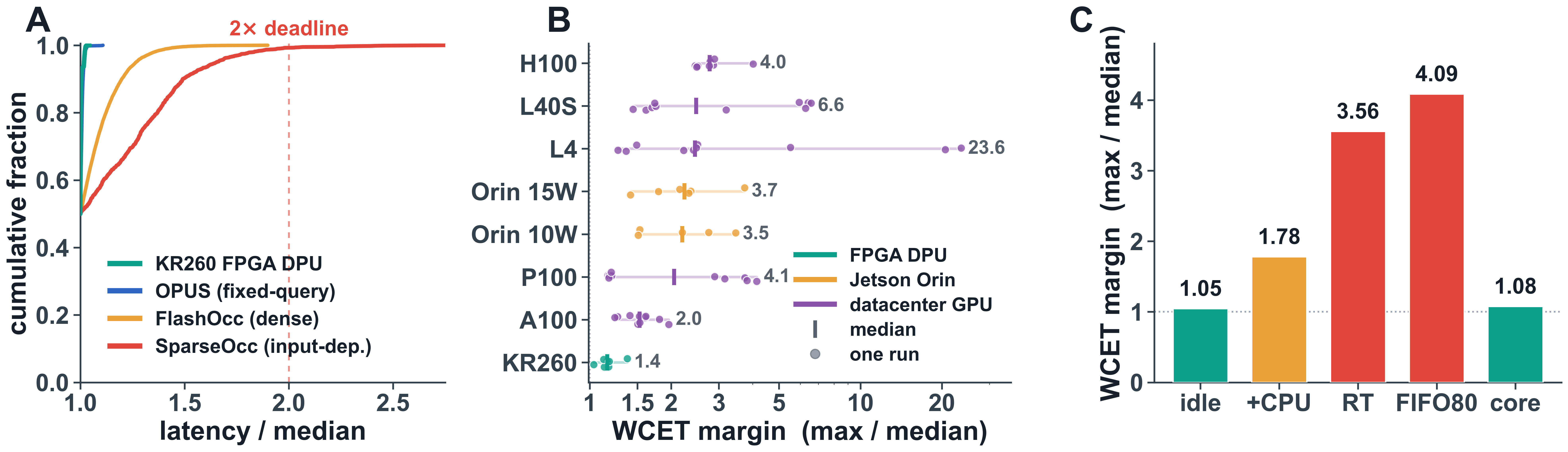}\hfill
\includegraphics[width=0.388\textwidth]{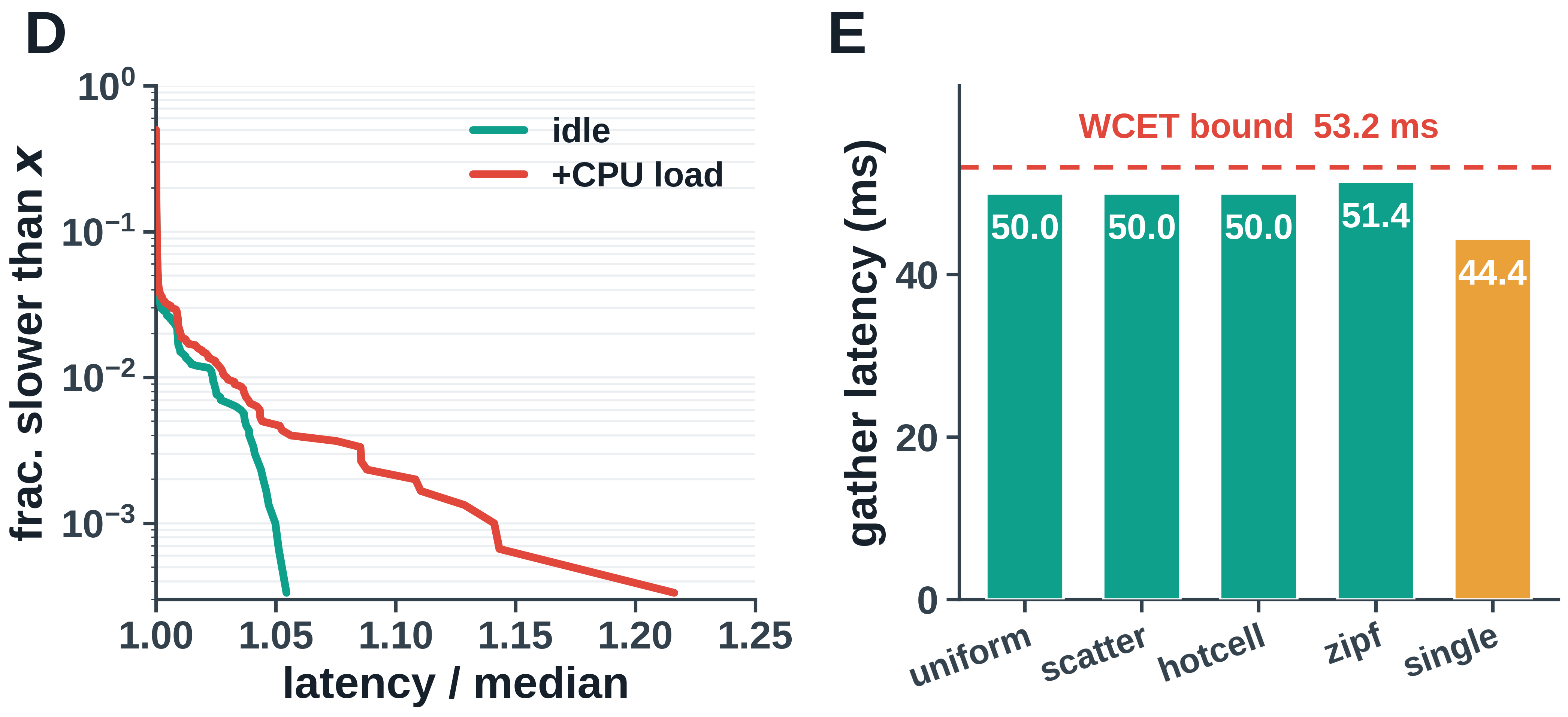}
\caption{Latency determinability decomposes into three measured legs (a--c), and the deployed gather IP is deterministic on the axes fixed work controls (d--e). \textbf{(a)} Per-frame latency CDF normalized by each series' median: fixed-work series (KR260 DPU, fixed-query OPUS) stay near-vertical, while dense FlashOcc and input-dependent SparseOcc grow tails, SparseOcc past the $2\times$ deadline; the discriminator is fixed work, not sparsity. \textbf{(b)} Per-run tail ratio (max/median) across the full hardware sweep over five seeds in idle and loaded conditions: the KR260 clusters tightly, while Jetson Orin and datacenter GPUs spread to large seed- and load-dependent worst cases. \textbf{(c)} KR260 host-path leg: real-time priority makes the polling DPU runner worse (3.56 per-thread, 4.09 whole-process FIFO80), while whole-process core dedication restores the tight envelope (1.08). \textbf{(d)} Median-normalized latency tail (CCDF) of the gather IP, idle versus compute-only CPU load: the median is fixed by cycle count, so only the host polling tail extends; memory-saturating load stretches the streaming window itself (text). \textbf{(e)} Five pathological point-to-pillar distributions at fixed $N_{\mathrm{points}}$: no frame exceeds 53.2\,ms at idle memory, and the degenerate single-pillar scene only runs faster.}
\label{fig:determinism}
\end{figure*}

\begin{figure*}[t]
\centering
\begin{subfigure}[t]{0.305\linewidth}
\centering
\includegraphics[width=\linewidth]{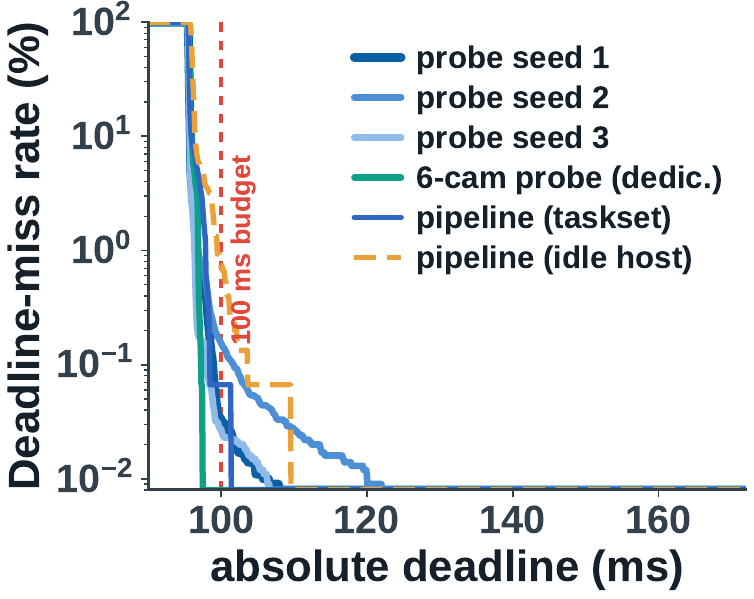}
\caption{}\label{fig:absdeadline}
\end{subfigure}\hfill
\begin{subfigure}[t]{0.305\linewidth}
\centering
\includegraphics[width=\linewidth]{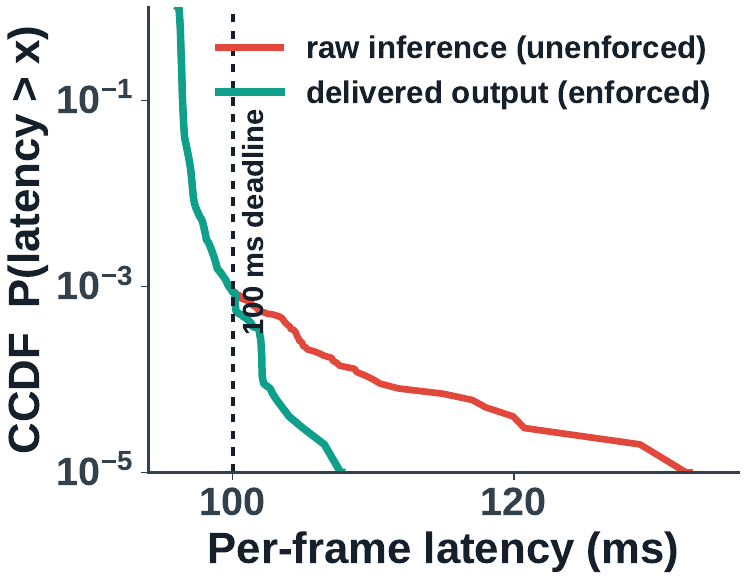}
\caption{}\label{fig:delivery}
\end{subfigure}\hfill
\begin{subfigure}[t]{0.305\linewidth}
\centering
\includegraphics[width=\linewidth]{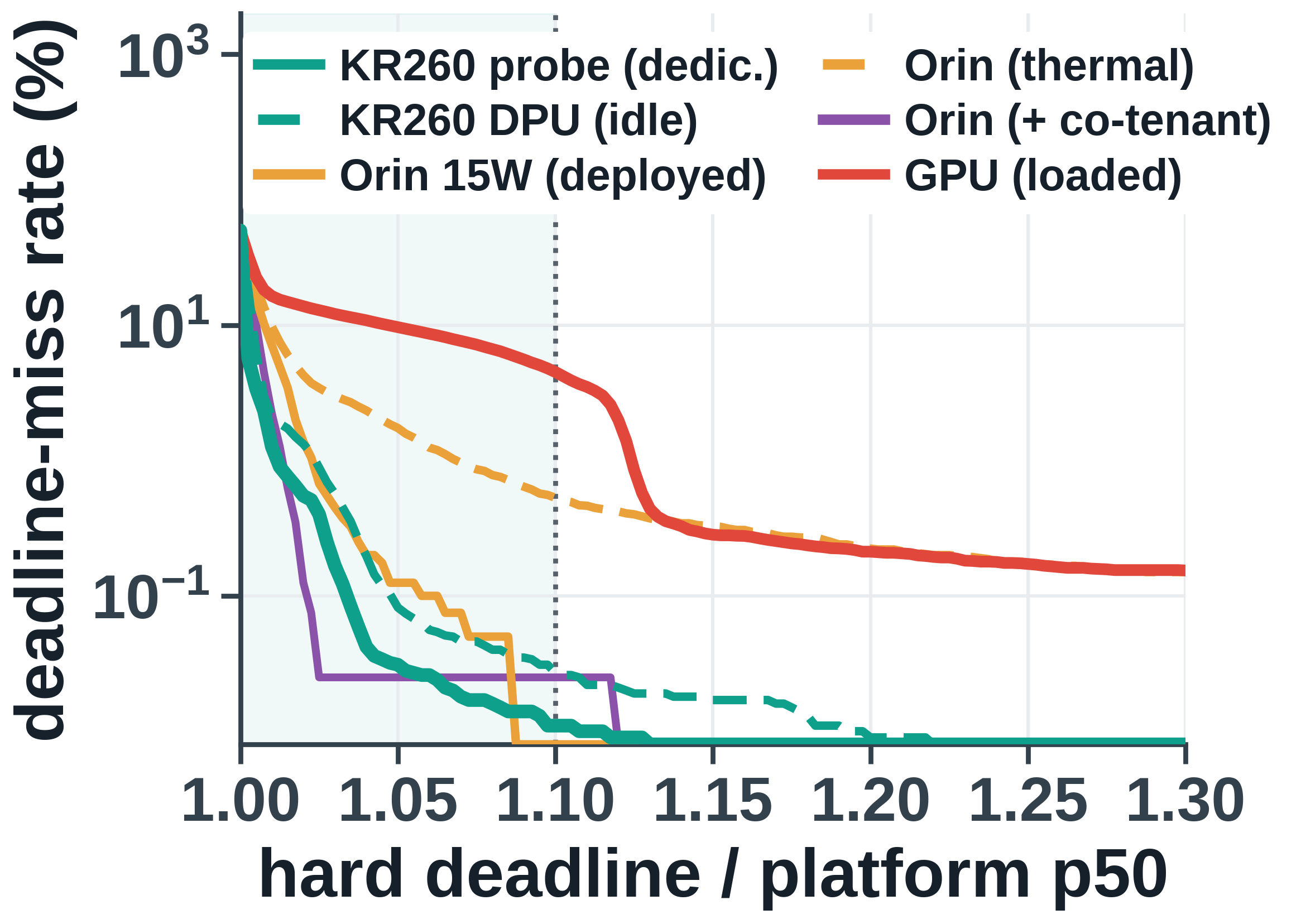}
\caption{}\label{fig:payoff}
\end{subfigure}
\caption{Deadline behavior at $100$k-frame scale. \textbf{(a)} Absolute-deadline view: the six-inference DPU probe, a fixed-work proxy rather than the full FlashOcc pipeline of \cref{tab:budget}, misses the $100$\,ms budget on $0.027$ to $0.152\%$ of frames across three dedicated seeds, maxima $124$ to $169$\,ms; shorter traces understate the tail. \textbf{(b)} Enforced delivery (CCDF): the raw tail ($132.5$\,ms max) crosses the budget on $0.088\%$ of frames, $2.75\times$ the unenforced seed-1 trace of (a), a probe effect of the enforcement timer's own host work; the delivered-output curve terminates at $107.8$\,ms, $68$ of the $88$ late frames as stale-flagged fallbacks and the rest publishing fresh inside the enforcer's wake window; two further enforced runs with tenfold-spanning raw miss rates deliver every frame by $107.6$ and $110.2$\,ms, the $10.2$\,ms worst excursion a host-thread wake tail a PL-side timer would remove. \textbf{(c)} Relative-deadline view, shaded band marking deadlines within a $10\%$ margin of each platform's median: the dedicated KR260 probe misses $0.011\%$ at that margin; a loaded GPU and a thermally-throttled Orin miss several percent at theirs.}
\label{fig:deadline}
\vspace{-12pt}
\end{figure*}

Throughout, \emph{host-visible latency} is wall time from job submission to completion on the host, the quantity a deadline check sees; a \emph{CPU burner} is a spin loop pinned to a named core; a \emph{co-tenant} is an unrelated GPU workload sharing the accelerator (steady: fixed-size matmul loop; bursty: randomized sizes and sleeps); \emph{seeds} are repeated runs launched at different times with freshly initialized processes, nothing else varying. The $100$k-frame runs replay a fixed input tensor, so work is input-invariant by construction and residual variation is platform-induced; \cref{fig:determinism}e varies content explicitly. \emph{Conservative occupancy} names the fallback of \cref{eq:watchdog}: the previous output ego-warped, flagged stale, unable to contain objects first visible in the missed frame. Latency probes use ResNet-50@224 INT8 on the DPU; accuracy runs use the deployed FlashOcc xmodels (trunk $256{\times}704$, BEV stage \texttt{bev\_clamp32}); BEV-stage latency traces use the structurally identical \texttt{bev\_reluc} compile. Software: XRT 2.13.0, Vitis AI 2.5/3.0, PYNQ-DPU, Ubuntu 22.04. All runs use batch $1$, one frame in flight: DPU INT8 on the stock B4096 overlay, custom IPs at $200$\,MHz, GPU probes FP16 at stock clocks, Orin TensorRT $10.3$ INT8 at the stated power mode. Latency curves time the current frame's inference only; fallback delivery is measured separately (\cref{fig:delivery}).

\Cref{fig:determinism}(a,b) show the platform gap. Pro6000 is tight when idle (max/p50 1.07) but reaches $2.13\times$ under a co-tenant; shared-A100 FlashOcc reaches $1.72\times$; Orin NX varies from $1.1\times$ to $4.4\times$ across power modes and seeds; a consumer RTX 4090 reaches $2.75\times$ at $100$k frames while idle (CV $7.4\%$), consistent with GPU timing studies~\cite{gpu-nondeterminism-gap,dart-rtss,rtgpu}. The KR260 probe tail is the smallest and grows slowest because accelerator work is fixed: $1.44\times$ idle and $2.0\times$ dedicated at $100$k frames, the rare host spikes behind these tails handled by the delivery-bounding watchdog.

The model sweep separates sparsity from determinability (\cref{fig:determinism}a). OPUS is sparse but fixed-query, with CV $1.1\%$ and max/p50 1.11. SparseOcc is sparse but input-dependent, with CV $27.5\%$, max/p50 2.75, and a $+102$\,ms tail. FlashOcc is dense and fixed-compute, with CV $12.6\%$ and max/p50 1.72 on the shared node. SparseOcc latency is uncorrelated with occupied voxels or object count (\texttt{corr} $=-0.02$ and $+0.01$), so scene size does not predict the tail.

Panel (b) adds the seed dimension: the five-seed spread caps Orin near $3.7\times$, the $4.4\times$ above being its separate 100k worst-seed run (\cref{fig:window}); a bursty co-tenant drives Pro6000 CV to $36.7\%$; on the KR260, CPU burners push max/p50 from 1.045 to 1.78. RT priority makes the polling DPU runner worse (3.56 per-thread, 4.09 whole-process \texttt{SCHED\_FIFO} priority 80), while whole-process core dedication reduces the loaded tail to 1.077 (\cref{fig:determinism}c); this is measured on the DPU submission runner, the DDR-streamed PL IP path characterized separately (\cref{sec:exp-ips}).

\subsection{KR260 hardware closure}
\label{sec:exp-ips}
\label{sec:exp-onboard}

Both custom IPs have cycle-bounded, input-invariant compute (\cref{tab:hlsips}). Each targets the real K26, passes C/RTL cosimulation, and closes timing at 200\,MHz. The INT16 datapath is bit-exact against its own integer simulation; against floating-point references the IPs are numerically matched: gather to FlashOcc CUDA \texttt{bev\_pool\_v2} within max-absolute error $2.6\mathrm{e}{-5}$, resize to torch bilinear \texttt{align\_corners=True} within $1\mathrm{e}{-13}$.

\begin{table}[!h]
\centering
\caption{Custom PL IPs synthesized with Vitis HLS on the KR260 K26. Both have cycle-bounded compute, match their references numerically at full rig size, pass C/RTL cosimulation, and close timing at 200\,MHz. Compute WCET is the cycle count at the $200$\,MHz deployment clock, not at Fmax.}
\label{tab:hlsips}
\small
\setlength{\tabcolsep}{4pt}
\renewcommand{\arraystretch}{1.0}
\begin{tabular*}{\columnwidth}{@{}L{0.25\columnwidth}@{\extracolsep{\fill}}L{0.68\columnwidth}@{}}
\toprule
\multicolumn{2}{@{}l}{\textbf{Gather IP} \hfill LSS view transform, FlashOcc BEVPoolv2} \\
\midrule
Exactness & max-abs $2.6\mathrm{e}{-5}$ vs CUDA reference \\
Fixed work & $N_{\mathrm{points}}{=}302{,}558$ static point visits \\
Kernel / timing & ${\approx}0.93$M cycles, ${\approx}4.7$\,ms; Fmax 211\,MHz \\
Post-route use & DSP 15\%, BRAM 10\%, LUT 5\% \\
\midrule
\multicolumn{2}{@{}l}{\textbf{Resize IP} \hfill BEV-neck bilinear upsample} \\
\midrule
Exactness & max-abs $1\mathrm{e}{-13}$ vs torch reference \\
Fixed work & fixed output grid and interpolation taps \\
Kernel / timing & $688{,}023$ cycles, $3.44$\,ms; Fmax 230\,MHz \\
Post-route use & DSP 10\%, BRAM 54\%, LUT 9\% \\
\bottomrule
\end{tabular*}
\end{table}

In the full-rig deployment, large static tables stream from DDR; this changes latency but not the loop bounds. Host-side input flush costs $0.11$\,ms and output readback $0.17$\,ms (p50), so the $52.5$\,ms median is the \texttt{ap\_start}-to-\texttt{ap\_done} kernel window: PL compute plus DDR streaming. Compute-only CPU burners leave the median fixed and lengthen only the host polling tail (\cref{fig:determinism}d); deliberate memory saturation on the other three cores stretches the window itself to $65.1$\,ms p50 and $88.7$\,ms max over $2000$ frames, because streaming shares DDR bandwidth. The per-frame traffic is a compile-time byte ledger ($8.44$\,MB: $2.51$ rank tables, $0.17$ intervals, $0.64$ activation streams, $5.12$ output), but dividing it by bulk-copy bandwidth underestimates the window since rank-indexed reads are random-access; we therefore report the compute cycle count ($4.7$\,ms) as the analytical component and the memory-saturated envelope as the operative ceiling. Holding $N_{\mathrm{points}}$ fixed across five pathological point-to-pillar distributions, the median stays within $[44.4, 51.4]$\,ms and no frame exceeded $53.2$\,ms at idle memory (\cref{fig:determinism}e; a later $2000$-frame idle trace peaks at $54.3$\,ms, the value the budget table carries); the degenerate single-pillar scene only runs faster. Resize is likewise stable at p50 $19.67$\,ms, CV $0.20\%$, its gap to the $3.44$\,ms cycle count again DDR streaming. The unified B4096-DPU bitstream closes timing with $+0.15$\,ns slack at 65\% LUT, 59\% FF, 69\% BRAM, 83\% DSP.

\begin{table}[!h]
\centering
\caption{Composed per-frame budget of the full-resolution deployment from per-stage board measurements (serial, no overlap, from the trunk input; host-side decode/normalization excluded). Evidence classes: cycle-bounded (analytic), measured (idle envelope), stressed (deliberate memory saturation).}
\label{tab:budget}
\footnotesize
\setlength{\tabcolsep}{3pt}
\renewcommand{\arraystretch}{1.0}
\begin{tabular*}{\columnwidth}{@{}L{0.47\columnwidth}@{\extracolsep{\fill}}rrl@{}}
\toprule
Stage & p50 & worst & evidence \\
\midrule
Image trunk ($6\times$, $256{\times}704$) & $6{\times}45.5$ & $6{\times}49.8$ & measured \\
Gather IP kernel (compute) & \multicolumn{2}{c}{$4.7$} & cycle-bounded \\
Gather kernel incl.\ DDR streaming & $52.5$ & $54.3$ & measured \\
\quad under memory saturation & $65.1$ & $88.7$ & stressed \\
BEV DPU subgraphs ($3$) & $399.3$ & $408.2$ & measured \\
Resize IP ($2$ calls) & $39.3$ & $40.2$ & measured \\
Host marshaling (flush, readback) & $0.3$ & $1.5$ & measured \\
CPU head (predictor MLP, NumPy) & $6730$ & $6780$ & measured \\
\quad on DPU ($1{\times}1$ convs, LUT softplus) & $505$ & $534$ & measured \\
\midrule
Accelerated stages, serial & $\approx 764$ & $\approx 803$ & composed \\
\quad with DPU head & $\approx 1269$ & $\approx 1337$ & composed \\
\bottomrule
\end{tabular*}
\end{table}

\subsection{Accuracy and deadline behavior}
\label{sec:int8-accuracy}
\label{sec:exp-int8}
\label{sec:exp-payoff}
\label{sec:exp-budget}

The relevant target is the INT8 algorithm, because the DPU cannot run FP32. Quantizing only the BEV stage (clamp 32, trunk FP32) reaches $98\%$ of FP32 mIoU (32.19 to 31.6 on 6019 frames), isolating the clip cost. Reference configurations matter: an idealized clamp-16 reference gives 29.08; the reference matched to the deployed configuration (clamp 32, view transform quantized as deployed, INT16 at $1/128$ resolution) scores 31.43, and the board reaches 29.88 against it, a 1.55 gap attributable to the real per-layer quantizer realization. Quantizing the view transform to INT8 instead would cost 9.4 mIoU (22.01 vs 31.43), the direct justification for the INT16 gather datapath. The feature cosine 0.873: board trunk output and the quantizer's CPU simulation score identically against FP32 features (0.87323 both), so the deploy chain adds nothing on top of quantization itself; the number measures PTQ cost, not deployment error, whose task-level effect \cref{tab:accuracy} reports.

\begin{table}[!h]
\centering
\caption{INT8 deployment accuracy on Occ3D-nuScenes. Columns 256/1024/2048/full: nested first-$N$ validation subsets, showing consistency across evaluation scale. The idealized clamp-16 reference is a configuration mismatch kept for context; the matched reference uses the deployed configuration. FP32 is an unreachable DPU upper bound.}
\label{tab:accuracy}
\footnotesize
\setlength{\tabcolsep}{3pt}
\renewcommand{\arraystretch}{1.0}
\begin{tabular*}{\columnwidth}{@{}L{0.36\columnwidth}@{\extracolsep{\fill}}ccccc@{}}
\toprule
\multicolumn{6}{@{}l}{\textbf{mIoU at validation scale}} \\
\midrule
Configuration & 256 & 1024 & 2048 & full & \%FP32 \\
\midrule
FP32 upper & 25.0 & 30.0 & 34.8 & 32.19 & 100.0 \\
INT8 sim.\ (idealized c16) & 23.1 & 27.5 & 31.5 & 29.08 & 90.3 \\
INT8 sim.\ (matched c32) & -- & -- & -- & 31.43 & 97.6 \\
\textbf{Board INT8} & \textbf{23.8} & \textbf{28.1} & \textbf{32.1} & \textbf{29.88} & \textbf{92.8} \\
\midrule
\multicolumn{6}{@{}l}{\textbf{VRU IoU on full validation set}} \\
\midrule
Configuration & bicyc. & motorc. & pedest. & \multicolumn{2}{c}{note} \\
\midrule
FP32 upper & 11.5 & 14.4 & 16.8 & \multicolumn{2}{c}{no DPU path} \\
INT8 sim. & 7.4 & 9.1 & 12.2 & \multicolumn{2}{c}{algorithm ref.} \\
\textbf{Board INT8} & 5.5 & \textbf{11.1} & \textbf{12.9} & \multicolumn{2}{c}{this work} \\
\bottomrule
\end{tabular*}
\end{table}

The board exceeds the idealized reference at every evaluated scale; pedestrian and motorcycle IoU exceed it too, while bicycle drops most (\cref{tab:accuracy}). \Cref{fig:occ} shows the board output following the INT8 reference and preserving visible VRU structure.

We evaluate deadlines in two frames. The relative frame sets each hard deadline as a fixed margin over the platform median, testing whether a platform reliably meets a deadline sized to its own throughput: at a $10\%$ margin over $100$k frames the dedicated KR260 probe (one ResNet-50@224 inference per frame) misses $0.011\%$, against \textbf{4.5\%} for the loaded GPU and \textbf{0.5\%} for the thermally-throttled Orin at their own margins (\cref{fig:payoff}); shorter traces can show zero misses at this margin, so measured maxima are reported as envelopes, not bounds. The absolute frame scores the deployment against a fixed $100$\,ms perception budget: at urban $50$\,km/h the vehicle travels $1.4$\,m per budget interval. On this budget, the six-inference DPU frame-budget probe (six sequential ResNet-50@224 INT8 inferences per frame, $6\times$ the relative-frame probe's accelerator work, held fixed; exact configuration in the released code and traces) runs at $95.5$\,ms median and misses on $0.027$ to $0.152\%$ of unenforced frames across three $100$k-frame seeds ($0.070\%$ pooled), with per-seed maxima of $124.3$, $133.2$, and $169.0$\,ms; the enforced run of \cref{fig:delivery} is a separate trace, on which the watchdog converts every miss into a flagged fallback, every output landing within $7.8$\,ms of the deadline over $100$k enforced frames ($10.2$\,ms worst over all three enforced runs). The full-resolution FlashOcc pipeline is far heavier than this proxy (\cref{tab:budget}): accelerated stages compose to roughly $0.8$\,s per frame serial, and the CPU head, a pointwise two-layer MLP over the $200{\times}200$ grid ($22$\,GFLOP), measures $6.7$\,s as an unoptimized FP32 NumPy reference on the A53 cores; recompiled as two $1{\times}1$-convolution DPU subgraphs with an int8 lookup-table softplus bridge it runs at $505$\,ms median (stage medians $89.6$, $389.3$, $31.7$\,ms; independent medians need not sum to the end-to-end median; $97.3\%$ argmax agreement with the FP32 head), $13\times$ faster than that unoptimized reference. The full pipeline thus composes to about $1.3$\,s per frame. Input downscaling does not buy this back: the BEV stage is independent of image resolution, and halving the trunk input without retraining collapses accuracy ($4.32$ vs $20.48$ mIoU paired, under a superseded view-transform scale, so the ratio is the evidence), and halving the BEV grid needs an architecture-consistent neck. A $10$\,Hz tier therefore requires a retrained reduced-grid model, a measured engineering boundary rather than a knob.

\subsection{Robustness, calibration, and the application context}
\label{sec:exp-robust}

On-board full-val mIoU is $29.88$ with a $95\%$ bootstrap CI of
$[29.61, 30.15]$ at the frame level and $[28.58, 31.11]$ scene-clustered; board-versus-reference
comparisons are paired on identical frames. Splitting by scene description gives day $30.26$ ($4449$ frames), rain $29.80$ ($1088$), night $\mathbf{16.76}$ ($602$; rain and night overlap on rainy nights). Rain costs almost nothing; night halves accuracy. The clip threshold is not the sensitive part: sweeping $C \in \{16, 24, 32, 48, 64\}$ on the night subset moves mIoU by at most $0.7$ ($16.4$ at $C{=}16$ vs $15.7$ at the deployed $C{=}32$), so the clipping recipe is robust to the lighting shift; night loss traces to the backbone.

Injecting camera-artifact perturbations before the backbone (matched INT8; baseline $28.9$ on the same first-$1000$-frame subset, differing from the full-set value): Gaussian blur $\sigma{=}1/2/4$ gives
$13.8/7.9/3.5$ mIoU; exposure gain $0.5\times$/$2\times$ gives $25.2/23.4$; sensor noise gives
$18.2$. Blur dominates, halving accuracy at one pixel of spread; moderate exposure change is nearly free; none of these can change the amount of work on the fixed-work path.

Calibration drift is probed by perturbing the rig pitch on all cameras (stale static tables
relative to the images): $0.1/0.2/0.5$ degrees cost nothing measurable and $1.0$ degree costs
$0.1$ mIoU against the $28.9$ baseline; tables regenerate offline on recalibration.

As for where the remaining accuracy loss lives: of all bicycle ground-truth voxels, the board
predicts $49.2\%$ free and $29.5\%$ background surfaces, keeps $5.7\%$ as bicycle, and scatters the remainder across driveable surface ($6.5\%$), vegetation, and
small movable-class confusions: thin structures are erased toward free space, a recall
problem quantization amplifies (on-board VRU recall: pedestrian $14.8\%$, motorcycle $12.3\%$,
bicycle $5.7\%$; occupied-any $43.8\%$ mean).

The fallback is priced directly: ego-warping the previous on-board prediction by one and two $2$\,Hz keyframes ($500$\,ms steps, five times the staleness a $100$\,ms miss produces, conservative upper bounds forced by the dataset's keyframe rate): pedestrian recall falls $14.8 \to 11.6 \to 8.1\%$, bicycle $5.7 \to 4.3 \to 3.8\%$, occupied-any recall on pedestrian voxels $42.5 \to 39.7 \to 37.0\%$. The expected delivered pedestrian recall under the enforced $100$\,ms budget is $R \approx (1-p_{\mathrm{miss}}) R_{\mathrm{fresh}} + p_{\mathrm{miss}} R_{\mathrm{stale}}$ with $p_{\mathrm{miss}} = 8.8{\times}10^{-4}$: a degradation of about $0.003$ percentage points in expectation, under an independence assumption between spike timing and scene content.

Measured power, from INA260 board telemetry: KR260 $4.33$\,W idle, $5.8$ to $6.6$\,W under
inference, $8.7$\,W peak; Orin NX at its $15$\,W mode draws $7.4$\,W under INT8 inference
(VDD\_IN); the workstation GPU idles at $16$\,W and draws $329$\,W median under a co-tenant
(nvidia-smi). Reported for context, not as a contribution.

At matched full-model workload the shared A100 runs FlashOcc at $33.5$\,ms median, $57.5$\,ms maximum over $2$k frames, meeting the $100$\,ms budget on that trace and missing $0.5\%$ at $50$\,ms; a $2$k-frame trace is the finite-window trap of \cref{fig:window}, with no enforcement layer behind it.

%% file: 05-discussion.tex
\section{Discussion, Limitations, and Conclusion}
\label{sec:discussion}

This work is a deployment result, not a new occupancy backbone: a camera occupancy network becomes easier to certify when algorithm, accelerator mapping, and runtime policy all expose fixed work. The latency guarantee is layered, and only one layer is a bound: PL compute is cycle-bounded analytically; inference latency is a measured envelope whose maxima are sample-size-dependent; delivery is structurally guaranteed, possibly with a stale-flagged fallback. Raw miss rates span tenfold across enforced runs and raw maxima wander $125.2$ to $139.8$\,ms, yet every frame delivered within the deadline plus the envelope of \cref{sec:exp-payoff}. That envelope is itself measured: the enforcement timer is a host thread, and the PL-side timer that would remove its wake tail is not implemented. A back-to-back run after prior host activity misses more often than clean boot ($0.671$ vs $0.070\%$), drift that runtime monitoring must catch. The timing-closed K26 bitstream co-places the DPU and both resize IP instances; the gather IP runs from its own overlay, so the full pipeline has not run as one bitstream. The evidence is dataset replay on hardware, not a moving-vehicle test: on-car synchronization, planner-interface timing, actuation, and real-traffic fallback remain future validation, with runtime fault monitoring~\cite{antonante2023monitoring,schotschneider2025runtimemon,wang2024sotifsurvey}.

On-board VRU recall (pedestrian $14.8\%$, bicycle $5.7\%$) and night mIoU ($16.76$) are far below what a braking function needs, and the fallback lowers recall further on late frames (\cref{sec:exp-robust}); we claim no recall level as sufficient, since it comes from the system safety case~\cite{wang2024sotifsurvey}. The timing layer is model-independent, so a more accurate fixed-work model inherits it unchanged. The platform comparison is not matched in precision or isolation: GPUs run FP16, datacenter runs share cluster nodes, and the KR260 is dedicated. A real-time-configured GPU, Orin's DLA, or another FPGA would be fairer baselines.

Determinism does not remove compute cost (\cref{sec:exp-payoff}), but the condition is not FlashOcc-specific: fixed rigs and tensor shapes turn LSS splatting and known-shape resampling into static-index gathers with certifiable trip counts, extending FPGA determinism evidence~\cite{spade,embeddedunetfpga} to camera occupancy. The HLS sources, deployment recipe, and raw $100$k-frame traces are released at \href{https://github.com/Que121/falcon-kr260-release}{\texttt{github.com/Que121/falcon-kr260-release}}.